\documentclass[10pt,a4paper]{article}
\usepackage[a4paper,margin=0.78in,top=0.70in,bottom=0.72in]{geometry}
\usepackage{newtxtext,newtxmath}
\usepackage[T1]{fontenc}
\usepackage{microtype,xcolor,graphicx,booktabs,tabularx,array,amsmath,siunitx,enumitem,caption,subcaption,float,fancyhdr,titlesec,tcolorbox,tikz,hyperref,url,longtable,multirow,listings,lastpage}
\usetikzlibrary{positioning}
\definecolor{onebitteal}{RGB}{0,150,150}
\definecolor{onebitblue}{RGB}{54,82,112}
\definecolor{onebityellow}{RGB}{250,245,218}
\definecolor{onebitlight}{RGB}{238,248,248}
\hypersetup{colorlinks=true,linkcolor=onebitblue,citecolor=onebitblue,urlcolor=onebitblue}
\setlist[itemize]{leftmargin=1.4em,itemsep=2pt}
\setlist[enumerate]{leftmargin=1.6em,itemsep=2pt}
\titleformat{\section}{\Large\bfseries\color{onebitblue}}{\thesection}{0.6em}{}[\color{onebitteal}\titlerule]
\titleformat{\subsection}{\large\bfseries\color{onebitblue}}{\thesubsection}{0.6em}{}
\titlespacing*{\section}{0pt}{14pt}{7pt}\titlespacing*{\subsection}{0pt}{10pt}{4pt}
\newcommand{\WModel}{W1.58A16}\newcommand{\Qwen}{Qwen3-8B}
\newcommand{\TernaryFootnote}{\footnote{Here ``ternary'' is engineering shorthand, not a claim of a fully native single-plane ternary network. The evaluated model is a layer-wise mixed/adaptive representation: most linear weights use one ternary plane, while a salience-selected subset uses a second plane plus per-group offsets/scales and GPTQ-style compensation. The measured effective weight-information budget is approximately 1.64 bits/weight; embeddings, the language-model head, normalization parameters, activations, and KV cache remain higher precision.}}
\newtcolorbox{takeaway}{colback=onebitlight,colframe=onebitteal,boxrule=0.5pt,left=8pt,right=8pt,top=6pt,bottom=6pt,arc=1mm}
\newtcolorbox{productbox}{colback=onebityellow,colframe=orange!65!black,boxrule=0.5pt,left=8pt,right=8pt,top=6pt,bottom=6pt,arc=1mm}
\begin{document}

\begin{center}
{\fontsize{23}{27}\selectfont\bfseries\color{onebitblue}
Scaling Post-Training Ternarization to Qwen3-8B\\[2pt]
Capability Retention, Reproduction, Lossless Packing, and Packed Execution\par}
\vspace{5pt}{\large\color{onebitteal}\bfseries OneBit AI -- Technical Research Report\par}
\vspace{5pt}
\begin{tcolorbox}[colback=onebitlight,colframe=onebitteal,boxrule=0.5pt,width=0.95\linewidth]\centering
\textbf{\Qwen{} FP16} $\rightarrow$ \textbf{KOTMS} $\rightarrow$ \textbf{E2M-ATQ} $\rightarrow$ \textbf{GPTQ} $\rightarrow$ \textbf{\WModel{}} $\rightarrow$ \textbf{lattice packing} $\rightarrow$ \textbf{packed kernel}
\end{tcolorbox}
\vspace{3pt}{\small A scale-up study of aggressive post-training low-bit conversion on a single RTX 5070.}\vspace{6pt}\\
Anirudh Malik \quad|\quad Poojith Devan \quad|\quad M Sparsh Mehra\\
{\small OneBit AI | Technical Research Report | September 2026}
\end{center}

\begin{takeaway}
\textbf{Executive takeaway.}
The Qwen3-8B conversion reproduces the published TWLA W1.58 reference to within 0.05 points on the reported zero-shot average (62.00 versus 62.05). The local FP16 baseline is nevertheless 7.9\% different from the published baseline, so absolute converted-model scores are safer than ratios.

Under a matched eight-task protocol, the 8B conversion retains 78.5\% chance-corrected capability versus 69.6\% for the 4B checkpoint. The 8B advantage is +8.9 points and is concentrated in knowledge-heavy tasks.

The adaptive representation is approximately 1.64 bits/weight, but the complete model is not a 1.64-bit-per-parameter checkpoint. Only linear projections are targeted; embeddings, LM head, norms, activations and KV cache remain higher precision. Lattice-aware serialization reduces the recorded 16.9~GB checkpoint to 8.24~GiB with essentially unchanged perplexity.

The project then demonstrates direct packed execution at 15.52 tokens/s in 7.35~GiB, replacing the earlier CPU-offloaded regime of roughly 1.48 tokens/s. This is not a universal FP16 speedup claim: the 8B FP16 model did not fit the 12~GB card. A prototype packed GEMV remains 4.6$\times$ slower than FP16 cuBLAS on one shape, identifying kernel optimization as the remaining systems bottleneck.
\end{takeaway}

\begin{abstract}
Ultra-low-bit language models promise reductions in storage and memory traffic, but a nominal ``1.58-bit'' label does not specify the deployed representation or its execution cost. We study a scale-up of an aggressive post-training conversion pipeline from Qwen3-4B to Qwen3-8B.

The conversion uses KOTMS rotation, E2M-ATQ adaptive ternarization, and GPTQ-style error compensation in a weight-only A16 configuration. We do not claim these algorithms as new. Our contribution is the end-to-end scale-up characterization: an external reproduction gate, matched 4B/8B capability analysis, cross-corpus perplexity, effective-bit accounting, lossless lattice-aware packing, and direct packed execution.

The 8B model reaches a three-corpus perplexity ratio of 1.361$\times$, with WikiText-2, C4 and PTB ratios of 1.318$\times$, 1.393$\times$ and 1.371$\times$. On eight zero-shot tasks at $n=500$, mean accuracy is 64.6\% versus 72.4\% for FP16, corresponding to 78.5\% chance-corrected retention and a 7.8-point absolute cost. The matched 4B run retains 69.6\%, yielding an 8.9-point 8B advantage.

The packed checkpoint is 8.24~GiB and preserves the recorded perplexity to measurement precision. Direct packed execution reaches 15.52 tokens/s in 7.35~GiB, while a preliminary packed GEMV remains slower than FP16 cuBLAS. The result is a validated scale-up baseline: model size improves robustness to aggressive post-training discretization, actual serialization is solved for the measured artifact, and direct execution is feasible, while broader seeds, calibration distributions and kernel optimization remain open.
\end{abstract}


\section{Introduction}
Transformer language models are expensive because both their parameter count and their dense numerical representation create storage and data-movement costs.\cite{vaswani2017,qwen3} Post-training quantization provides a way to reduce those costs without retraining an entire model.\cite{gptq,awq,omniquant}

A single ternary symbol from $\{-1,0,+1\}$ contains
\begin{equation}b_T=\log_2(3)\approx1.585\end{equation}
bits of information. This is the basis of the 1.58-bit terminology used by native ternary work.\cite{bitnet,bitnet158} But the information content of one symbol is not the storage requirement of a complete checkpoint. Adaptive residual planes, offsets, scales, masks, rotations, embeddings, heads and metadata all matter.

Capability is a separate question. Native low-bit models are trained around their low-bit representation, whereas PTQ asks how much of an existing pretrained model survives a large numerical perturbation.\cite{bitnet,bitnet158,gptq} Recent ternary PTQ work has explored structured trit planes and calibration-efficient conversion.\cite{ptqtp,catq,twla}

Execution is a third question. AWQ demonstrates that low-bit inference gains depend on hardware-aware packing and kernels, not just the numerical bit width.\cite{awq} SmoothQuant makes a parallel point for weight-activation quantization: the representation and execution system must be designed together.\cite{smoothquant}

This paper therefore treats the Qwen3-8B experiment as five coupled questions:
\begin{enumerate}
\item Can the pipeline reproduce an external published reference?
\item How much capability survives at 8B?
\item Does 8B retain more capability than 4B under a matched protocol?
\item Can the representation become an actually smaller artifact?
\item Can that packed artifact be executed directly?
\end{enumerate}

The underlying quantizer is prior art. The new evidence is empirical and system-level: validated 8B scale-up, de-confounded scaling, packing, direct execution, and reproducibility safeguards.

\begin{productbox}
\textbf{Product interpretation.} A compressed model becomes useful only when capability, serialization and execution line up. The 8B experiment now supplies evidence for all three layers, while exposing a measurable kernel-performance gap rather than hiding it.
\end{productbox}

\section{Related Work}
BitNet established a trainable 1-bit Transformer formulation, while BitNet b1.58 popularized ternary weights as a practical low-bit target.\cite{bitnet,bitnet158} The present study differs because Qwen3 is converted after pretraining.

GPTQ introduced approximate second-order PTQ; AWQ emphasized salient-weight protection and hardware-friendly weight-only inference; OmniQuant optimized quantization parameters in a training-free framework.\cite{gptq,awq,omniquant} These works motivate the use of calibration data and non-uniform sensitivity.

Ternary PTQ is a rapidly developing direction. PTQTP uses structured trit planes, CAT-Q targets calibration-efficient ternarization, and TWLA combines KOTMS and E2M-ATQ with an activation-aware extension.\cite{ptqtp,catq,twla} We use TWLA's weight-side components in an A16 configuration rather than claiming a new ternarization algorithm.

Other extreme-compression work such as AQLM demonstrates that sub-3-bit representation is possible through more general additive codebooks and carefully designed serialization.\cite{aqlm} This reinforces the paper's emphasis on distinguishing nominal numerical precision from the actual file format.

\begin{table}[H]\centering
\caption{Positioning of the study.}
\begin{tabularx}{\linewidth}{l l X X}\toprule
Work & Regime & Main idea & Relation here\\\midrule
BitNet & Native low-bit & Train with low-bit weights & Conceptual foundation\\
BitNet b1.58 & Native ternary & Ternary weights during training & 1.58-bit target\\
GPTQ & PTQ & Hessian-aware compensation & Error-compensation basis\\
AWQ & PTQ & Activation-aware saliency & Non-uniform importance\\
OmniQuant & PTQ & Optimized quantization parameters & Calibration context\\
PTQTP & Ternary PTQ & Structured trit planes & Related representation\\
CAT-Q & Ternary PTQ & Calibration-efficient conversion & Related PTQ\\
TWLA & Ternary PTQ & KOTMS + E2M-ATQ + ILA-AMP & Algorithmic basis\\
This work & System study & Qwen3-8B validation + packing + execution & Scale-up characterization\\
\bottomrule\end{tabularx}\end{table}

\section{Model and Representation}
The target is Qwen3-8B, a dense 8B-class Qwen3 model with 36 transformer layers, hidden size 4096 and intermediate size 12288.\cite{qwen3} The conversion targets 252 linear projections: attention $q,k,v,o$ and MLP gate/up/down.

Only these linear projections are converted. Embeddings, LM head, norms, activations and KV cache remain FP16. Thus \WModel{} is weight-only A16, not an all-tensor low-bit graph.

Whenever the converted model is called ``ternary'', the qualification in the footnote applies.\TernaryFootnote{} The effective weight-information budget is approximately 1.64 bits/weight.

The adaptive structure uses four disjoint salience masks with
\begin{equation}
\texttt{num\_mask}=4,\qquad \texttt{orders}=[2,2,1,1].
\end{equation}
A recorded projection has coverages 0.5\%, 3.1\%, 10.0\%, and 86.5\%. Hence 96.5\% of that measured block uses one plane, while 3.6\% receives a second plane.

A compact representation is
\begin{equation}
\widehat W=(\mu+\alpha_0T_0+\alpha_1T_1)\odot M,\quad T_i\in\{-1,0,+1\}.
\end{equation}
The second plane is allocated selectively rather than uniformly.

\section{Conversion Pipeline}
\subsection{KOTMS rotation}
KOTMS applies a structured orthogonal transformation intended to make weight distributions more amenable to ternary approximation while controlling activation outliers.\cite{twla} For the 8B run, 252 projections were processed and the measured rotation time was 35 minutes. Metadata checks were performed before the multi-hour quantization stage.

\subsection{E2M-ATQ}
For each masked block,
\begin{equation}
\mu=\operatorname{rowmean}(W),\qquad W_c=W-\mu,
\end{equation}
followed by a threshold
\begin{equation}
\Delta_0=0.75\operatorname{rowmean}(|W_c|).
\end{equation}
The first support is
\begin{equation}
T_{0,i}=\begin{cases}+1&W_{c,i}>\Delta_0\\-1&W_{c,i}<-\Delta_0\\0&\text{otherwise}\end{cases}
\end{equation}
and a least-squares scale is selected. The residual is quantized into a second support. Alternating refinement updates $\mu,\alpha_0,\alpha_1$ while the discrete supports remain fixed.

Because $\mu\neq0$, reconstructed FP16 tensors can contain many non-zero floating values. A zero-count heuristic is therefore not a valid ternary-provenance test.

\subsection{GPTQ compensation}
GPTQ-style compensation redistributes quantization error using calibration activations and approximate inverse-Hessian information.\cite{gptq} The 8B configuration used 64 samples, 2048-token sequences, seed 0 and \texttt{percdamp}=0.01. A16 activations were retained, so ILA-AMP was skipped.

\section{Experimental Setup}
\subsection{Hardware}
All primary measurements used one NVIDIA RTX 5070 with 12,227~MiB VRAM and approximately 31~GB system RAM under Windows. The project used Python 3.10.11, CUDA 12.8, PyTorch, Transformers, Datasets and lm-evaluation-harness. Local patches added per-layer checkpointing, streaming evaluation, a Windows VRAM cap and an exposed damping parameter.

\subsection{Calibration}
WikiText-2 is the calibration corpus for the published-method gate. The widest 8B projection creates a memory requirement that grows linearly with calibration samples. Recorded projections are 6.39~GiB at 64 samples, 9.58~GiB at 96, 12.78~GiB at 128 and 25.56~GiB at 256. The practical allowed budget was about 10.98~GiB, making 64 the reliable maximum.

\begin{figure}[H]\centering\includegraphics[width=0.84\linewidth]{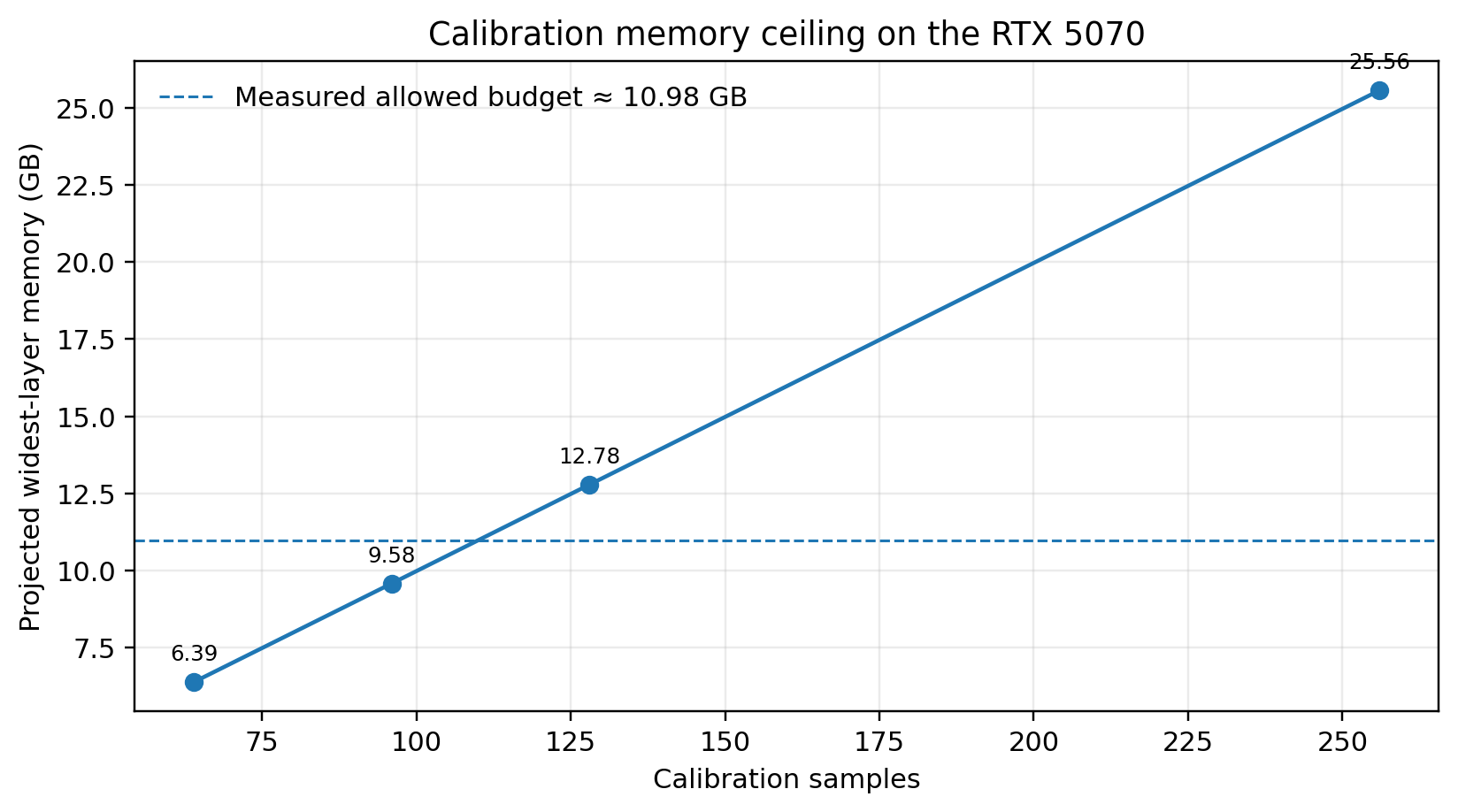}\caption{Calibration memory ceiling for the widest 8B projection.}\end{figure}

\subsection{Capability protocol}
Eight tasks were evaluated zero-shot at $n=500$: ARC-Challenge, ARC-Easy, BoolQ, HellaSwag, LAMBADA-openai, MMLU, PIQA and WinoGrande. ARC, HellaSwag and PIQA use normalized scoring consistent with the TWLA reference.

For fixed-choice tasks, chance-corrected retention is
\begin{equation}
R=\frac{A_s-B}{A_t-B},
\end{equation}
where $B$ is the larger of chance and majority-class floors. This avoids misleading raw ratios.

\subsection{Provenance safeguards}
Every benchmark requires zero missing/unexpected state keys, measurable weight changes and end-to-end perplexity reproduction. These checks were introduced because earlier project failures included wrong model identifiers, wrong checkpoint paths and malformed packed tensors that nevertheless produced plausible-looking numbers.

\section{Reproduction Gate}
Qwen3-8B is the TWLA reference model, so it enables an external gate rather than a purely internal consistency check.\cite{twla}

\begin{table}[H]\centering
\caption{Published TWLA reference versus our reproduction.}
\begin{tabular}{lrrr}\toprule
Metric & Published & Ours & Delta\\\midrule
FP16 WikiText-2 PPL & 9.00 & 9.715 & +7.9\%\\
\WModel{} WikiText-2 PPL & 12.52 & 12.758 & +1.9\%\\
FP16 zero-shot average & 69.09 & 69.57 & +0.7\%\\
\WModel{} zero-shot average & 62.05 & 62.00 & -0.08\%\\
\bottomrule\end{tabular}\end{table}

\begin{figure}[H]\centering\includegraphics[width=0.91\linewidth]{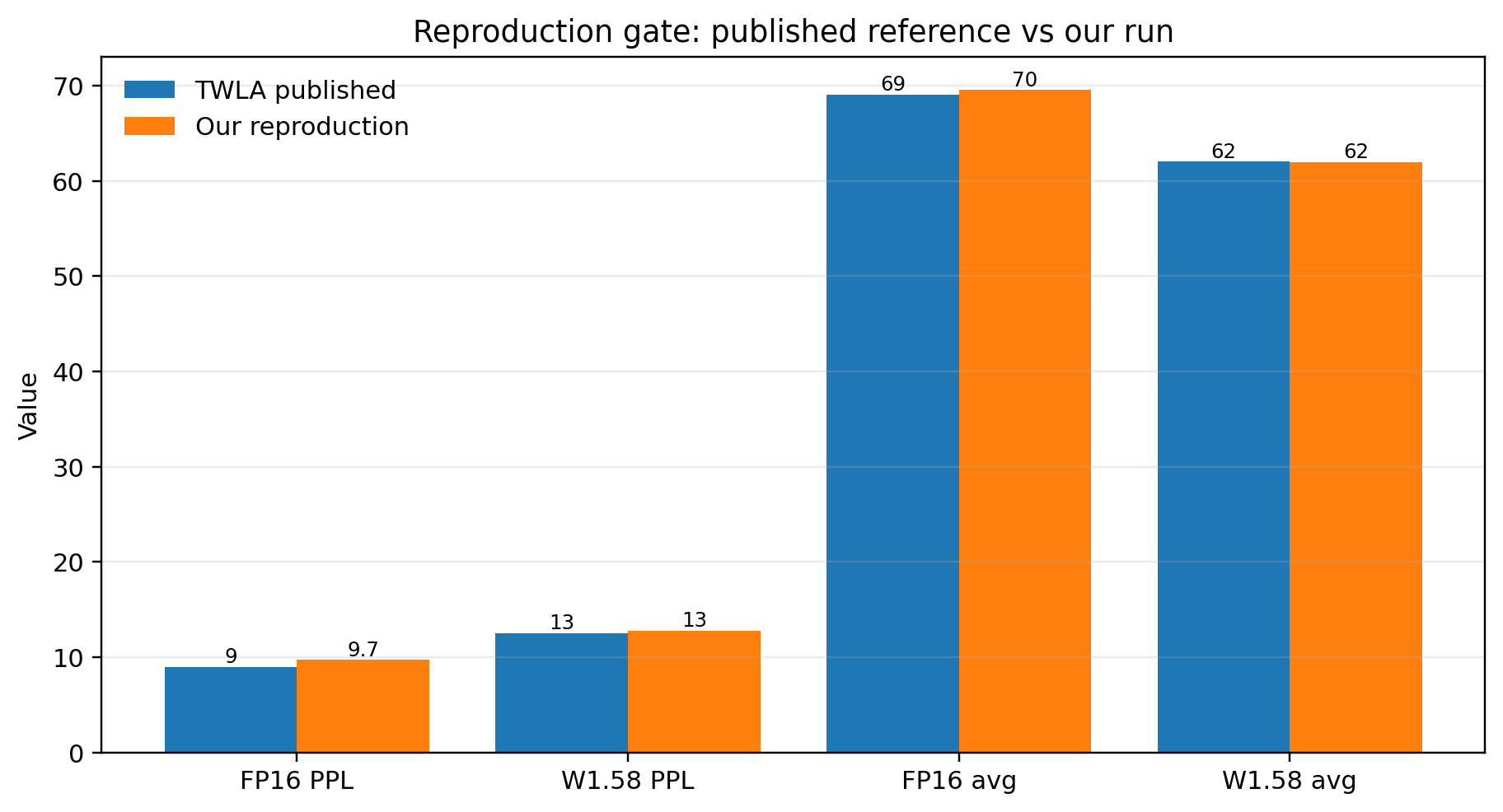}\caption{Reproduction gate. The converted-model result is within 0.05 points of the published reference.}\end{figure}

The gate passed. However, the local FP16 baseline differs from the published FP16 result by 7.9\%. This makes the local converted-to-FP16 ratio artificially favorable. The defensible statement is therefore that the converted absolute score is reproduced closely, not that our ratio improves on the published ratio.

The gate validates the integrated evaluation pipeline and provenance checks. It does not constitute an independent reimplementation of TWLA.

\section{Perplexity Results}
\begin{table}[h]\centering
\caption{Qwen3-8B perplexity.}
\begin{tabular}{lrrr}\toprule
Corpus & FP16 & \WModel{} & Ratio\\\midrule
WikiText-2 & 9.709 & 12.801 & 1.318$\times$\\
C4 & 15.275 & 21.271 & 1.393$\times$\\
PTB & 17.157 & 23.517 & 1.371$\times$\\
\midrule
Mean ratio & & & 1.361$\times$\\
\bottomrule\end{tabular}\end{table}

\begin{figure}[H]\centering\includegraphics[width=0.90\linewidth]{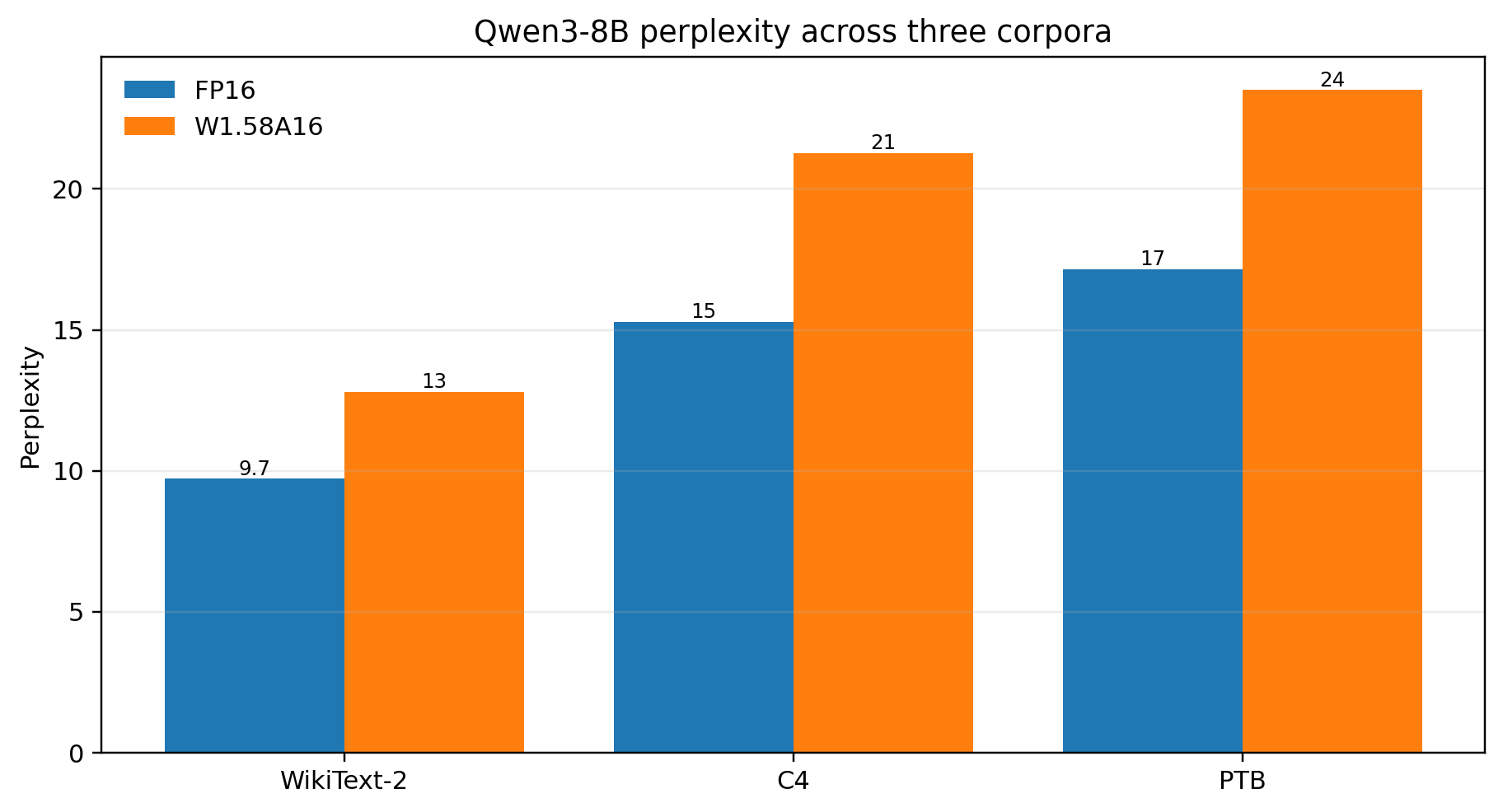}\caption{Perplexity across three corpora.}\end{figure}
\begin{figure}[H]\centering\includegraphics[width=0.78\linewidth]{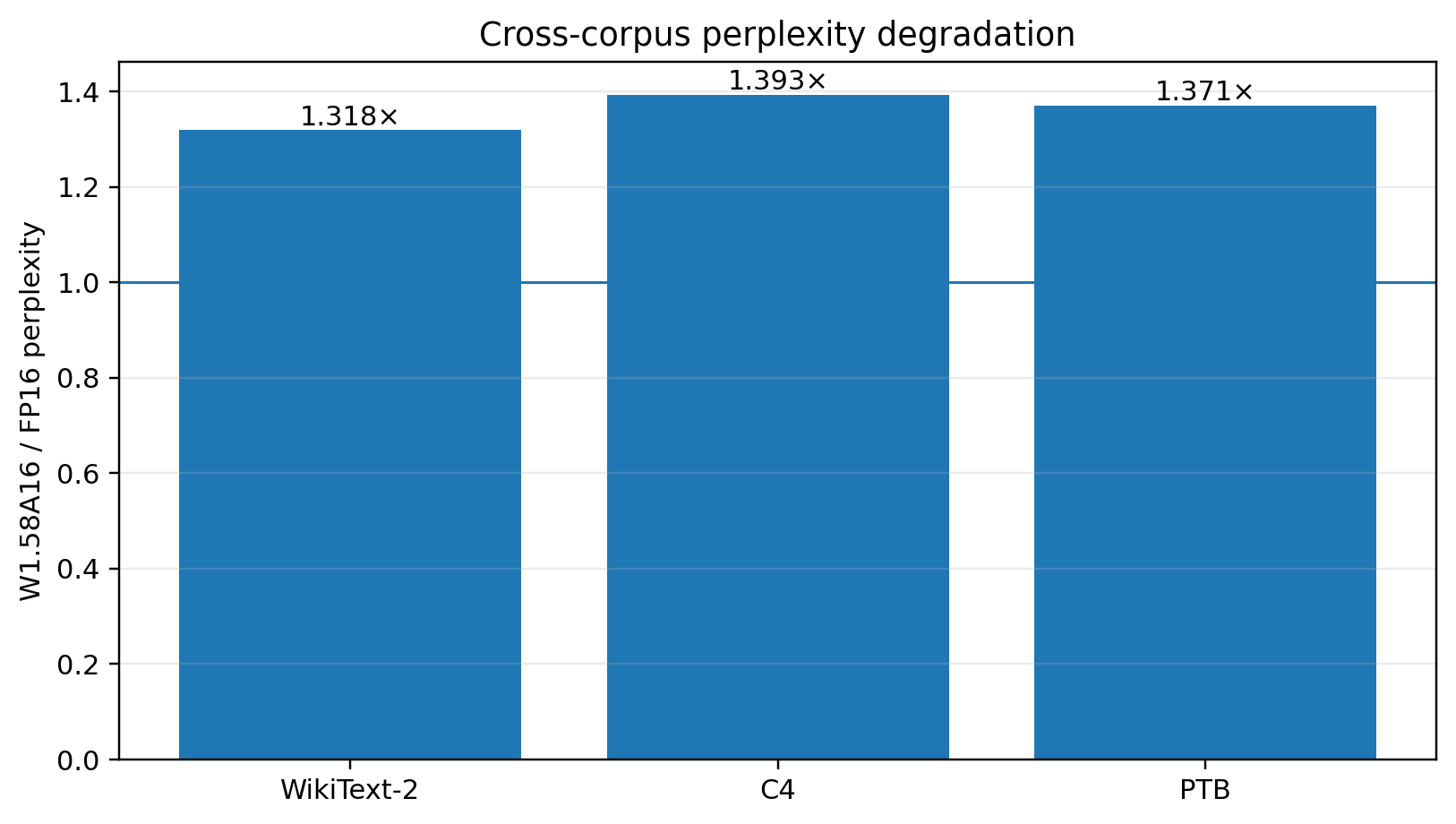}\caption{Relative perplexity cost across corpora.}\end{figure}

The 1.361$\times$ mean lies inside the pre-registered 1.25--1.45 range. WikiText-2 is the most favorable because it is also the calibration corpus. C4 is less favorable at 1.393$\times$, showing why one calibration-domain result is insufficient.

\section{Capability Results}
\begin{table}[H]\centering
\caption{Task-level capability.}
\scriptsize
\begin{tabular}{lrrrr}\toprule
Task & FP16 & \WModel{} & $\Delta$ & Retention\\\midrule
LAMBADA-openai & 66.8 & 60.6 & -6.2 & 90.7\%\\
BoolQ & 86.2 & 83.8 & -2.4 & 89.7\%\\
ARC-Easy & 81.6 & 72.8 & -8.8 & 83.9\%\\
WinoGrande & 68.6 & 65.0 & -3.6 & 80.6\%\\
HellaSwag & 64.4 & 56.0 & -8.4 & 77.9\%\\
PIQA & 80.2 & 72.6 & -7.6 & 74.3\%\\
MMLU & 75.55 & 61.0 & -14.55 & 71.2\%\\
ARC-Challenge & 55.8 & 45.0 & -10.8 & 60.0\%\\
\midrule
Mean & 72.4 & 64.6 & -7.8 & 78.5\%\\
\bottomrule\end{tabular}\end{table}

\begin{figure}[H]\centering\includegraphics[width=\linewidth]{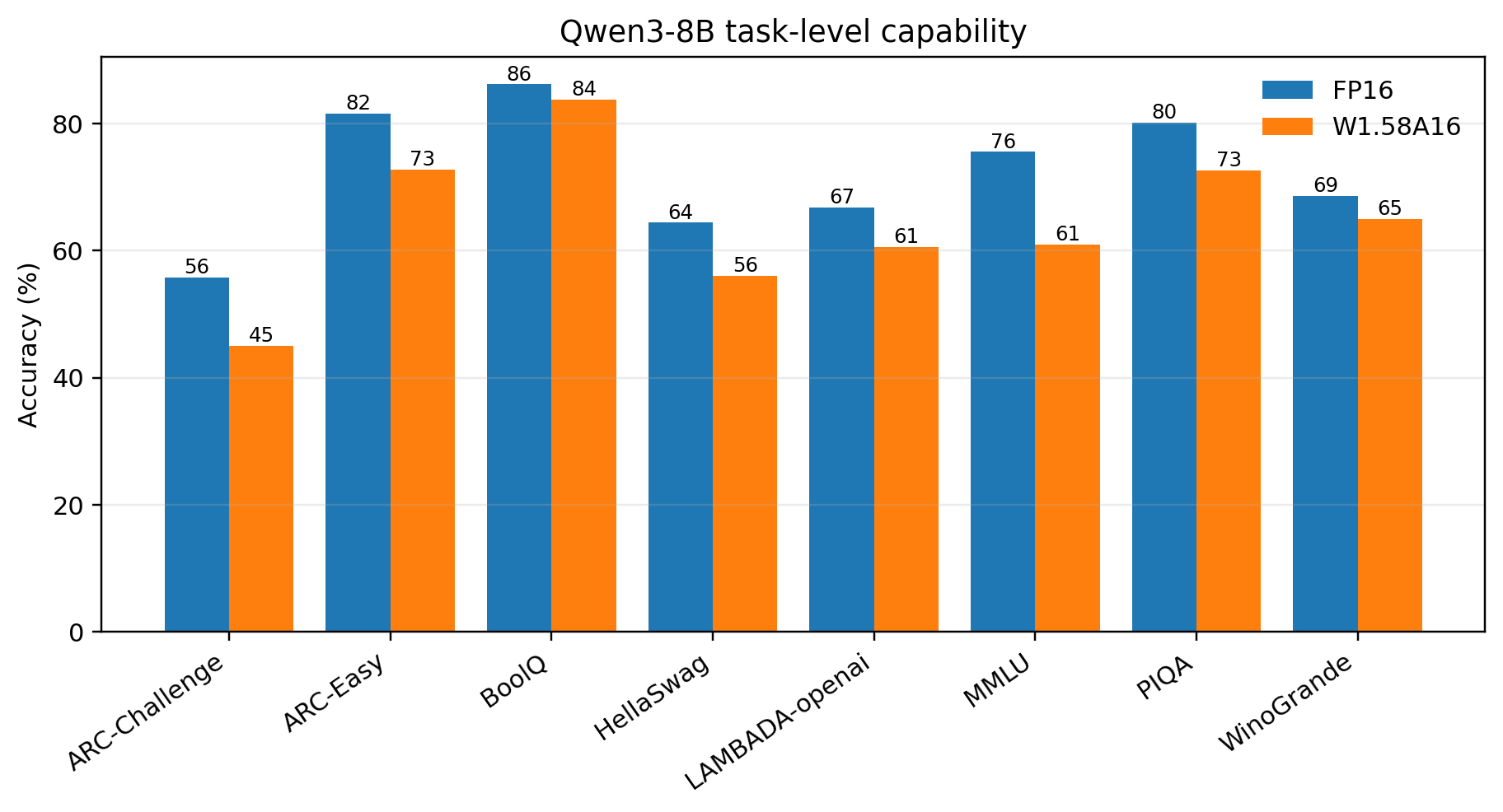}\caption{Task-level FP16 versus \WModel{} accuracy.}\end{figure}
\begin{figure}[H]\centering\includegraphics[width=0.92\linewidth]{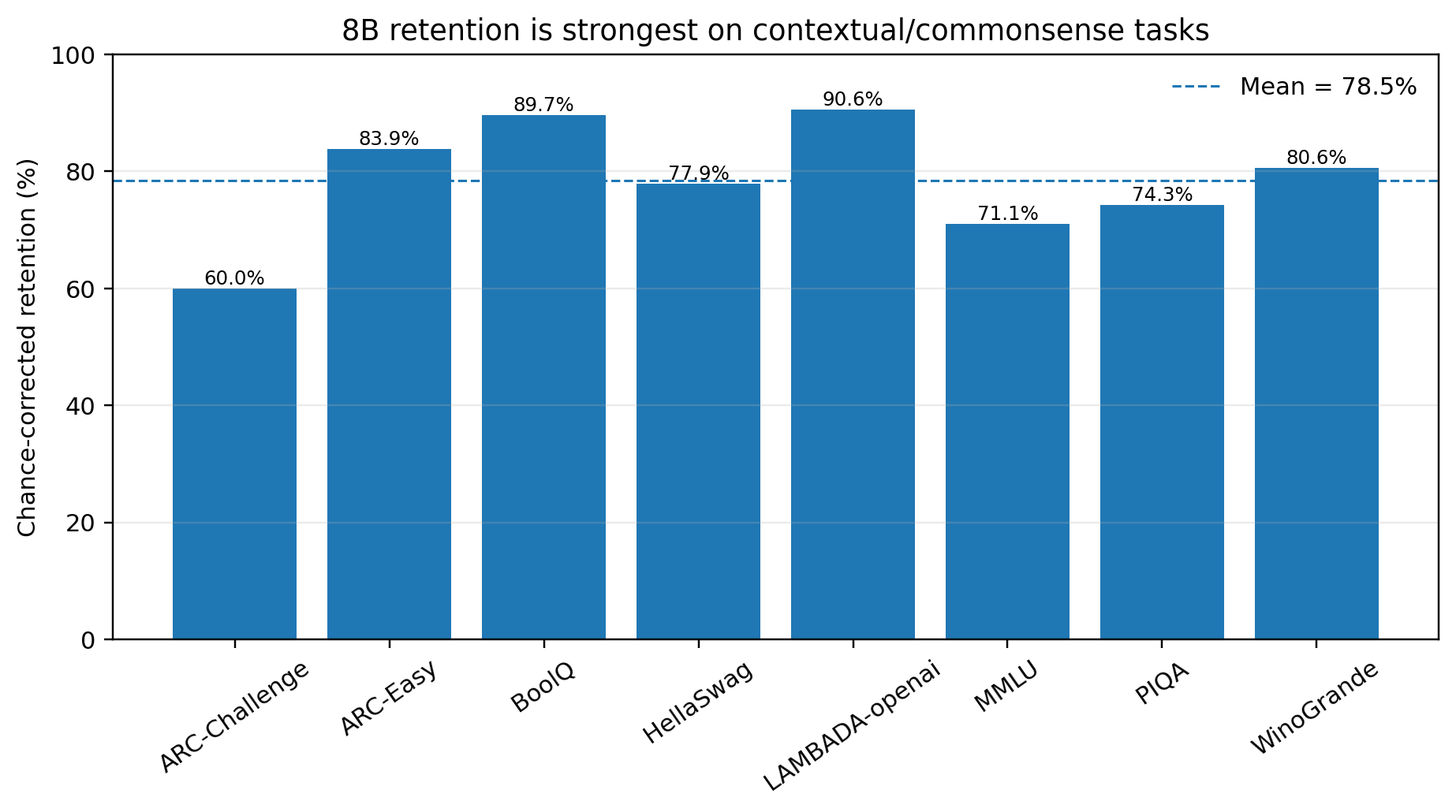}\caption{Chance-corrected retention.}\end{figure}

The mean accuracy cost is 7.8 points, but the loss is highly non-uniform. Contextual tasks remain relatively robust; knowledge-intensive tasks are weaker. This pattern is consistent with, but does not prove, a redundancy hypothesis.

\section{Mathematics: Degraded, Not Destroyed}
The pre-registered expectation was that mathematics would be at or below chance because several 4B mathematical tasks had previously approached chance. The 8B result contradicted that expectation.

\begin{table}[H]\centering
\caption{Reported Qwen3-8B mathematics-subject results.}
\begin{tabular}{lrrr}\toprule
Subject & FP16 & \WModel{} & Above 25\% chance?\\\midrule
Elementary mathematics & 70.1 & 54.2 & Yes\\
High-school statistics & 73.1 & 56.5 & Yes\\
College physics & 56.9 & 48.0 & Yes\\
High-school mathematics & 51.1 & 41.1 & Yes\\
Abstract algebra & 57.0 & 40.0 & Yes\\
College mathematics & 59.0 & 38.0 & Yes\\
\bottomrule\end{tabular}\end{table}

\begin{figure}[H]\centering\includegraphics[width=0.92\linewidth]{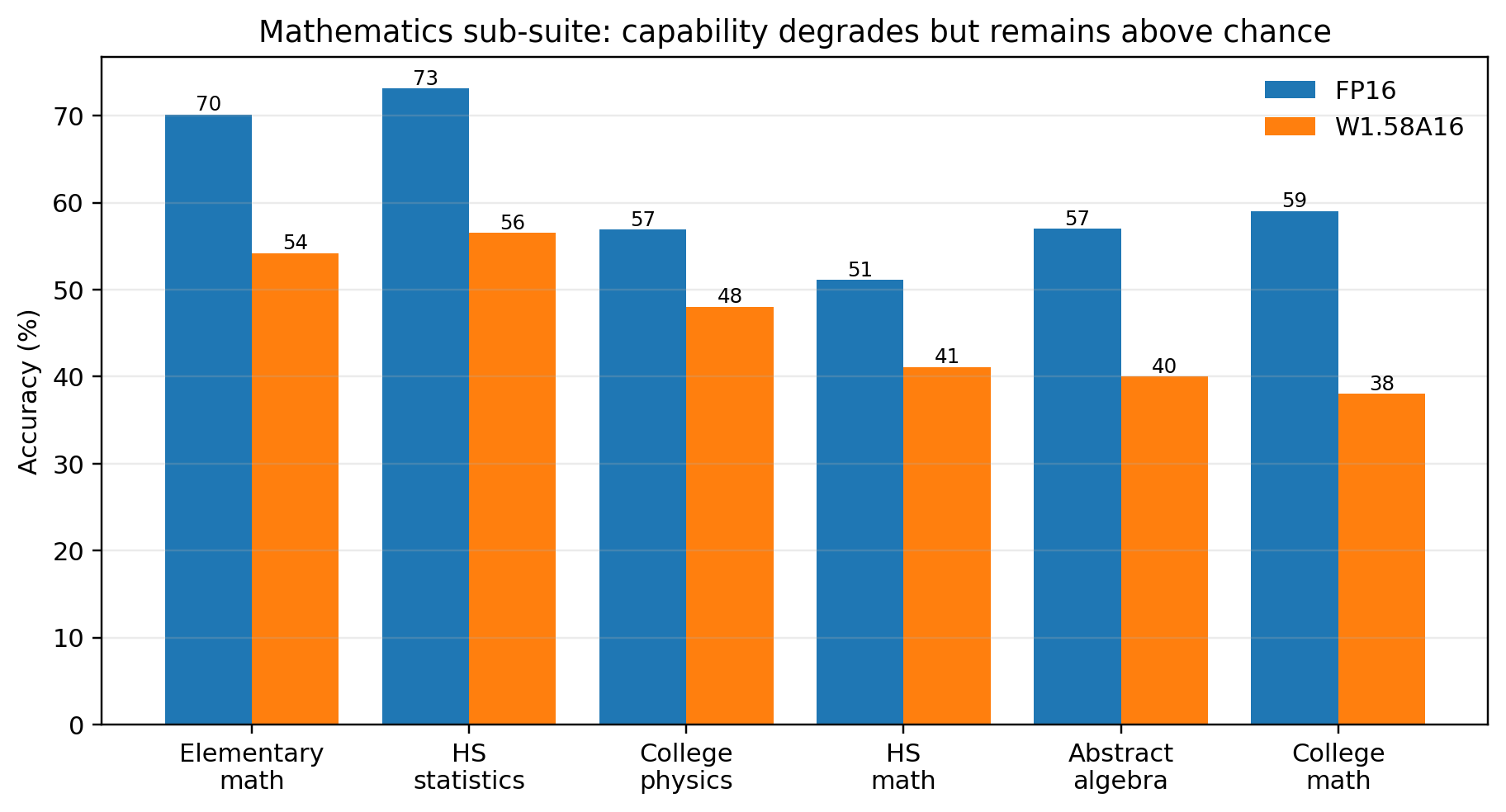}\caption{Mathematics-subject accuracy.}\end{figure}
\begin{figure}[H]\centering\includegraphics[width=0.80\linewidth]{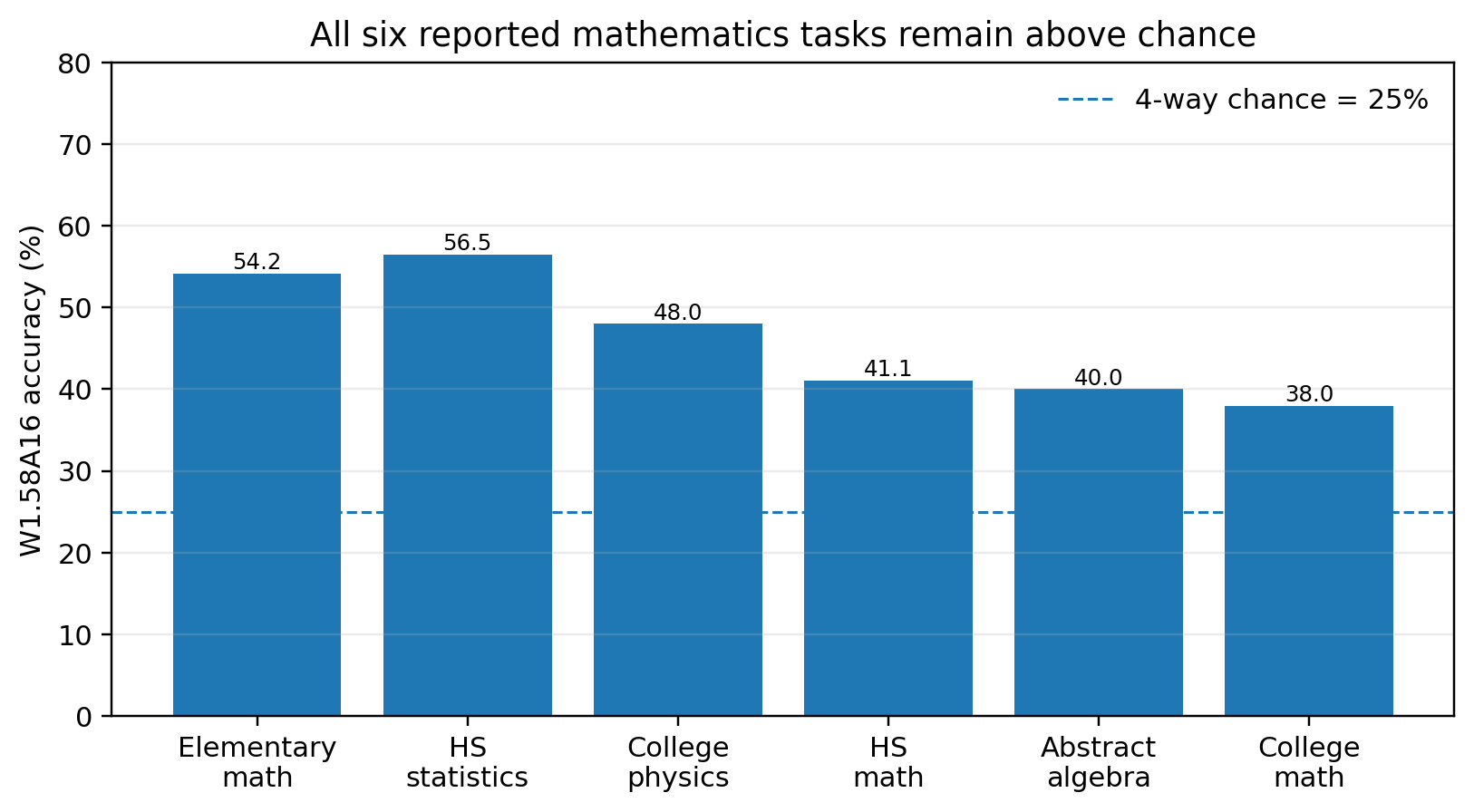}\caption{Every reported subject remains above the 25\% four-way chance floor.}\end{figure}

This is one of the strongest qualitative differences from the smaller model. The 8B conversion does not preserve full mathematical competence, but it preserves enough structure that all six reported subjects remain above chance.

This should not be called a scaling law. There are only two sizes, one seed, and the evaluation settings are finite. The correct conclusion is narrower: \textbf{in the matched scale-up study, increasing model size changes several low-bit failures from near-destruction to measurable residual capability.}

\section{Matched 4B--8B Scaling}
The original 4B/8B comparison was confounded. The 4B checkpoint was re-benchmarked at \texttt{percdamp}=0.01 and $n=500$ to match the 8B protocol. The resulting comparison isolates model size more cleanly.

\begin{table}[H]\centering
\caption{Matched-protocol chance-corrected retention.}
\scriptsize
\begin{tabular}{lrrr}\toprule
Task & 4B & 8B & 8B--4B\\\midrule
ARC-Challenge & 43.1 & 60.0 & +16.9\\
ARC-Easy & 54.7 & 83.9 & +29.2\\
BoolQ & 82.2 & 89.7 & +7.5\\
HellaSwag & 80.1 & 77.9 & -2.2\\
LAMBADA-openai & 72.4 & 90.6 & +18.2\\
MMLU & 60.3 & 71.1 & +10.8\\
PIQA & 82.8 & 74.3 & -8.5\\
WinoGrande & 81.6 & 80.6 & -1.0\\
\midrule
Mean & 69.6 & 78.5 & +8.9\\
\bottomrule\end{tabular}\end{table}

\begin{figure}[H]\centering\includegraphics[width=\linewidth]{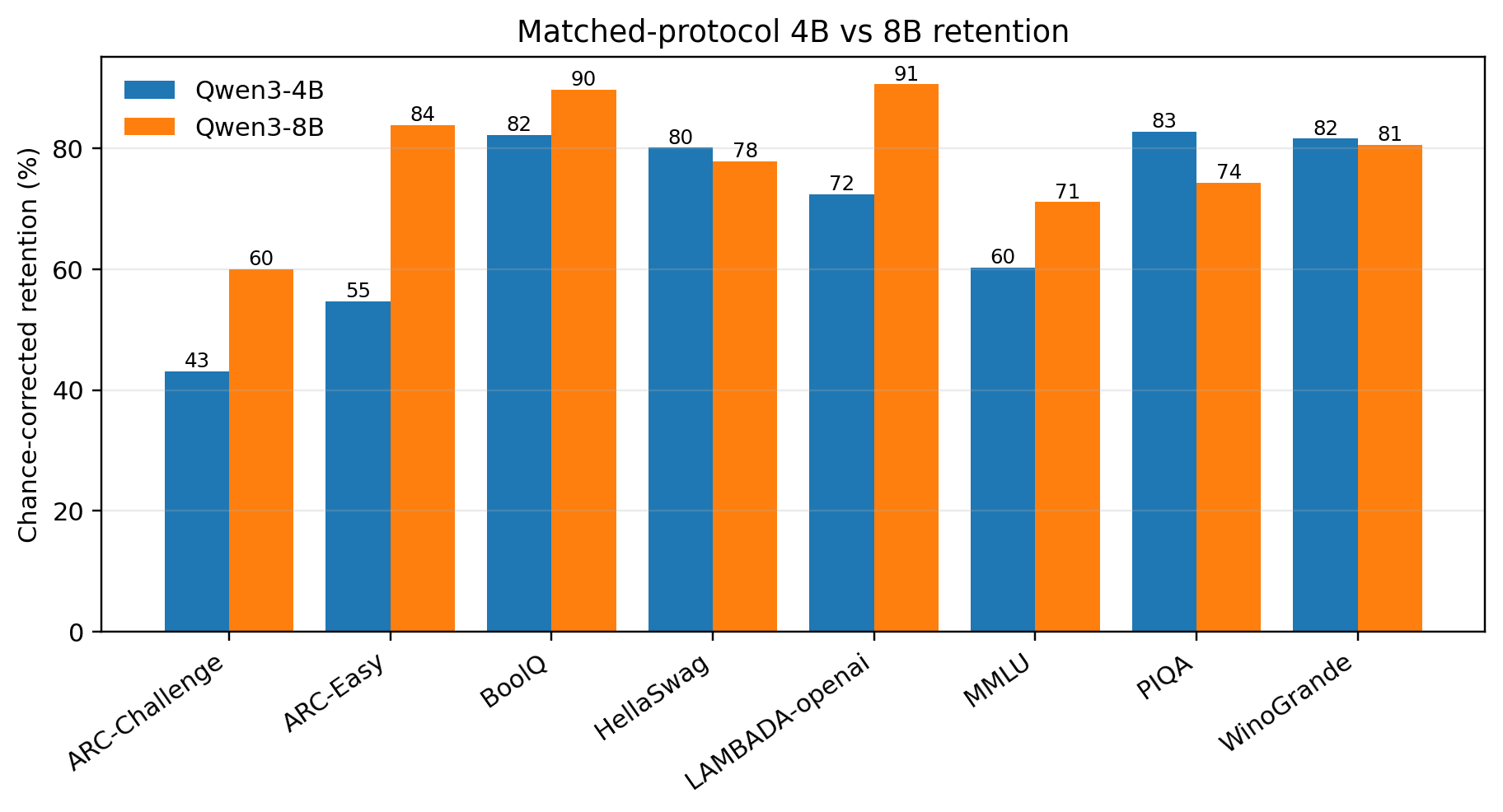}\caption{Matched-protocol 4B/8B retention.}\end{figure}
\begin{figure}[H]\centering\includegraphics[width=0.92\linewidth]{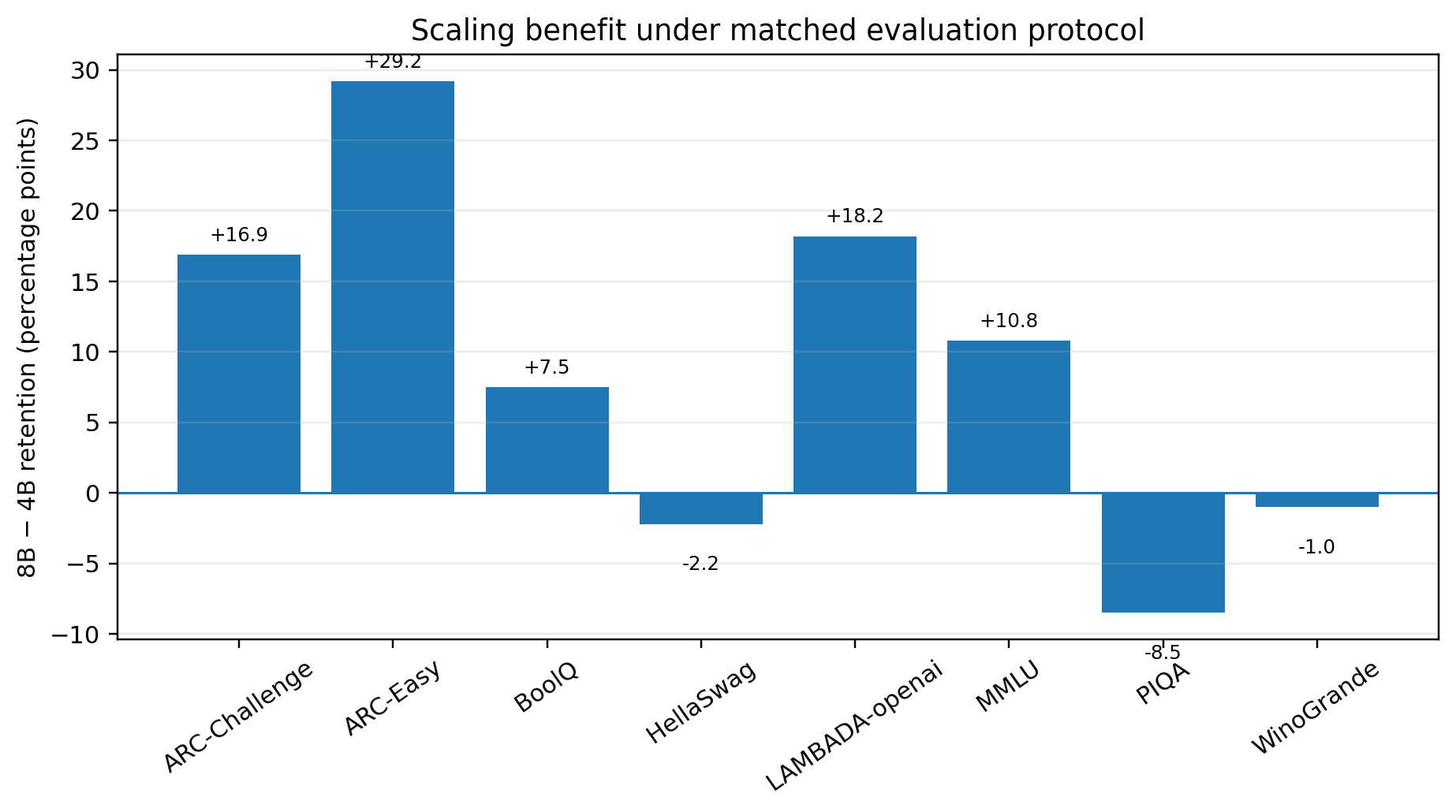}\caption{8B minus 4B retention.}\end{figure}

The 8B wins on six of eight tasks. The largest improvements are ARC-Easy (+29.2), LAMBADA (+18.2), ARC-Challenge (+16.9) and MMLU (+10.8). The two losses, HellaSwag and PIQA, are tasks where the 4B already retained high capability.

The pattern is consistent with redundancy: larger models may distribute useful information across more pathways, making aggressive rounding less likely to destroy a capability. Confirmation requires more scales and seeds.

\section{Representation Accounting}
The ideal single-plane ternary rate is 1.585 bits/weight. The measured adaptive weight representation is approximately 1.64 bits/weight. This is the number to use when discussing the quantized linear-weight representation, not the total checkpoint.

The representation can be viewed as
\begin{equation}
B_{\mathrm{eff}}\approx(1+f_2)\log_2(3),
\end{equation}
where $f_2$ is the fraction receiving a second plane. Salience grouping makes $f_2$ small.

The reconstructed FP16 tensor can contain many distinct values because the four masks have distinct offsets and scales. Therefore a naive zero-count or distinct-value test is insufficient. The project instead verifies state provenance and end-to-end perplexity.

The distinction between effective bit rate and checkpoint size is critical. The 8B artifact still contains embeddings, LM head, norms, rotation information and metadata. A model can have a near-ternary weight-information budget without occupying 1.64 bits per total parameter on disk.

\section{Packing: From Quantization to Serialization}
The first saved 8B checkpoint was fake-quantized: low-bit numerical values were stored in FP16 tensors. It therefore remained about 16.9~GB. The later lattice-aware format persists the discrete supports and associated parameters.

\begin{figure}[H]\centering\includegraphics[width=0.78\linewidth]{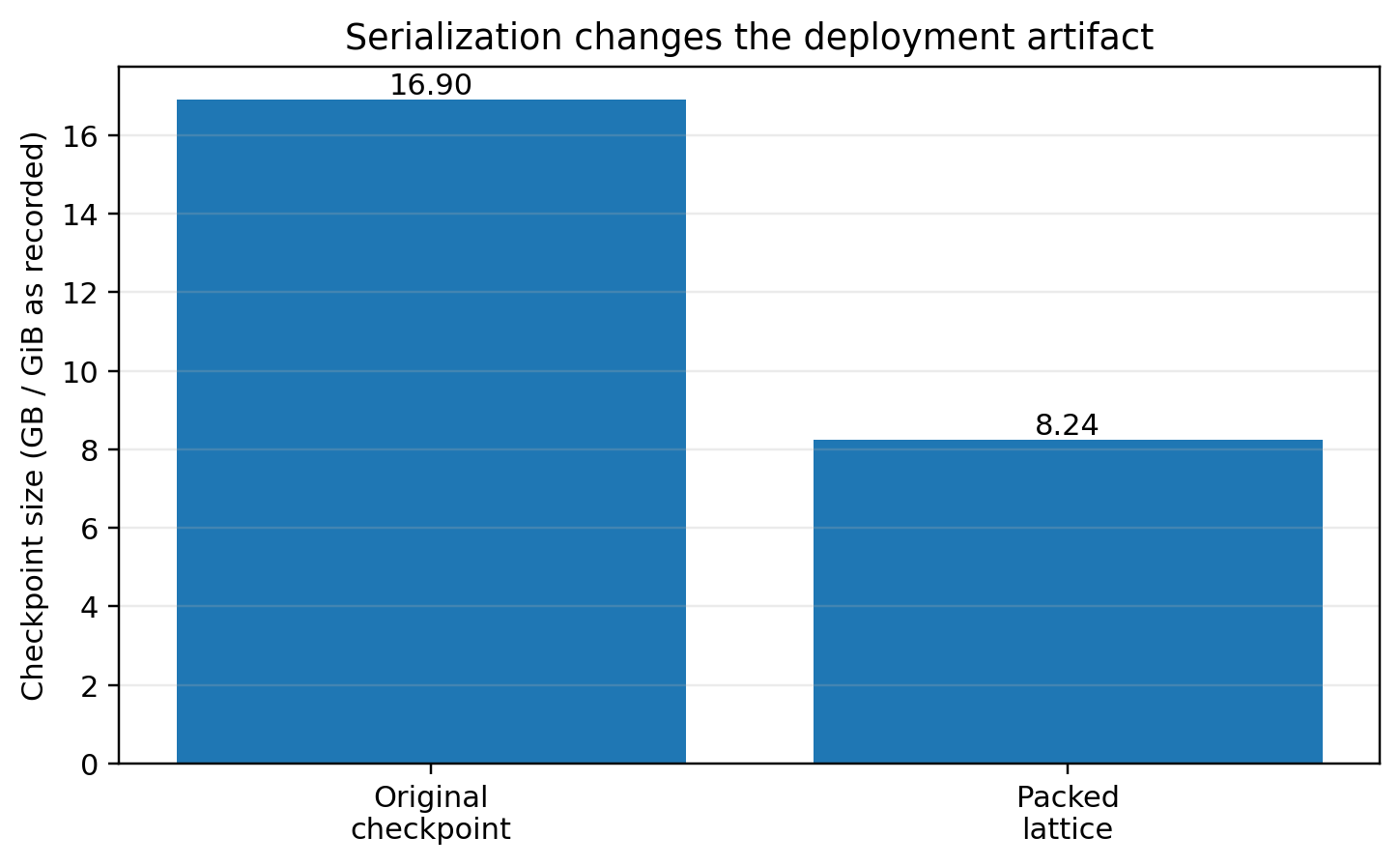}\caption{Recorded checkpoint size before and after lattice-aware packing.}\end{figure}

\begin{table}[H]\centering
\caption{Verified packed-artifact fidelity.}
\begin{tabular}{lrrr}\toprule
Metric & Original & Packed & Difference\\\midrule
Size & 16.9~GB & 8.24~GiB & --\\
WikiText-2 PPL & 12.758265 & 12.758180 & 0.000085\\
PTB PPL & 23.517294 & 23.516298 & 0.000996\\
\bottomrule\end{tabular}\end{table}

\begin{figure}[H]\centering\includegraphics[width=0.84\linewidth]{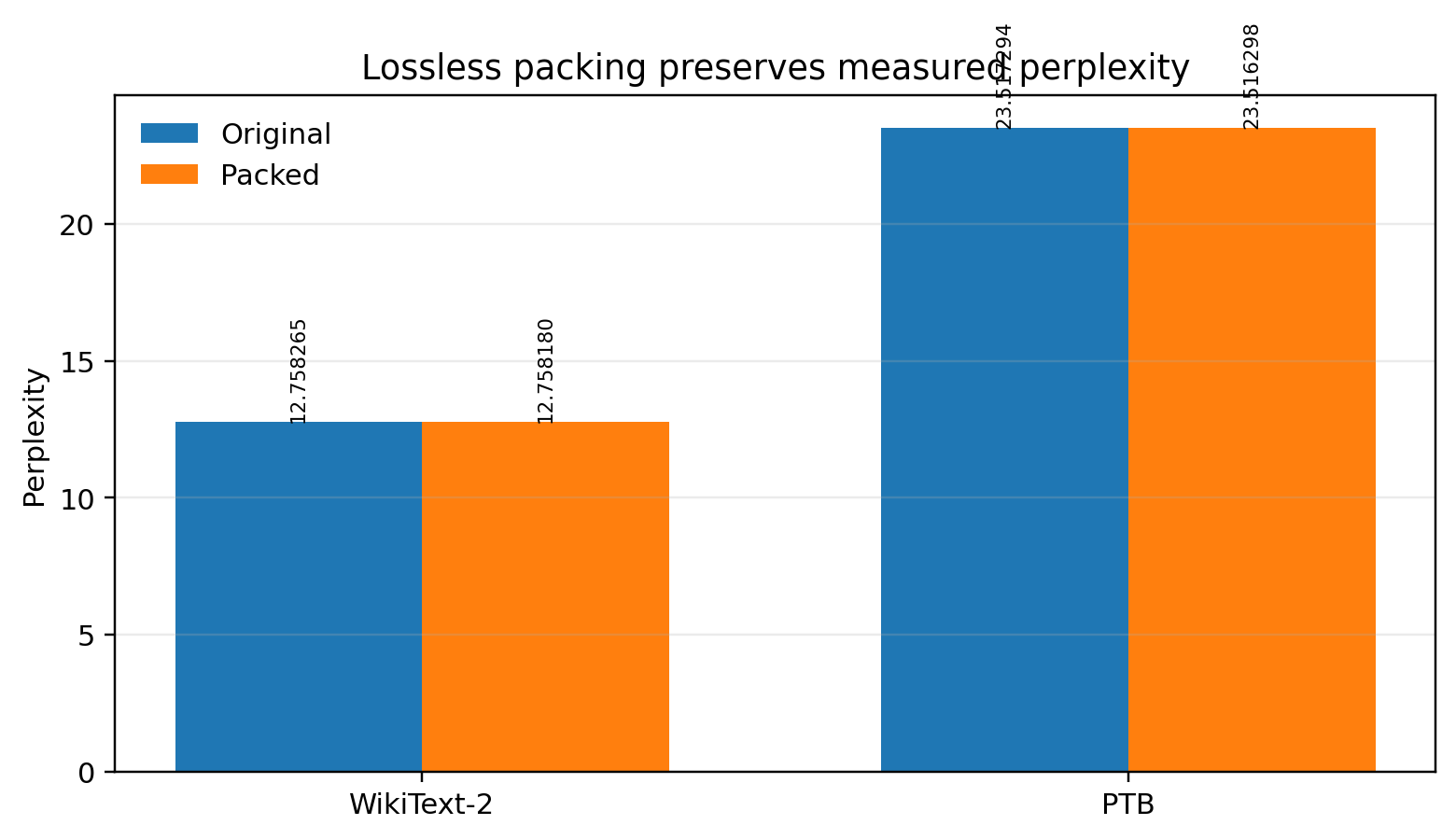}\caption{Perplexity preservation after packing.}\end{figure}

The WikiText-2 difference is approximately 0.0007\%. The source record reports approximately 0.036\% relative RMSE and about 101 seconds of packing time.

A post-hoc fitting attempt failed catastrophically: fitting a single simplified lattice to the saved FP16 tensor produced 50.5\% relative error and a perplexity of approximately 18,477. The reason is that the quantizer's discrete supports and activation-domain scales are not uniquely recoverable from the final floating reconstruction. The engineering lesson is direct: \textbf{serialization must be designed into the quantizer.}

\newpage
\section{Direct Packed Execution}
Before direct packed execution, the 8B model did not fit the 12~GB GPU and therefore relied on CPU offload. The recorded decode rate was about 1.5 tokens/s for both arms. Those measurements primarily captured PCIe transfer and offload overhead, not the intrinsic cost of low-bit arithmetic.

The updated packed runtime consumes the packed weights directly. It reaches 15.52 tokens/s in 7.35~GiB.

\begin{figure}[H]\centering\includegraphics[width=0.80\linewidth]{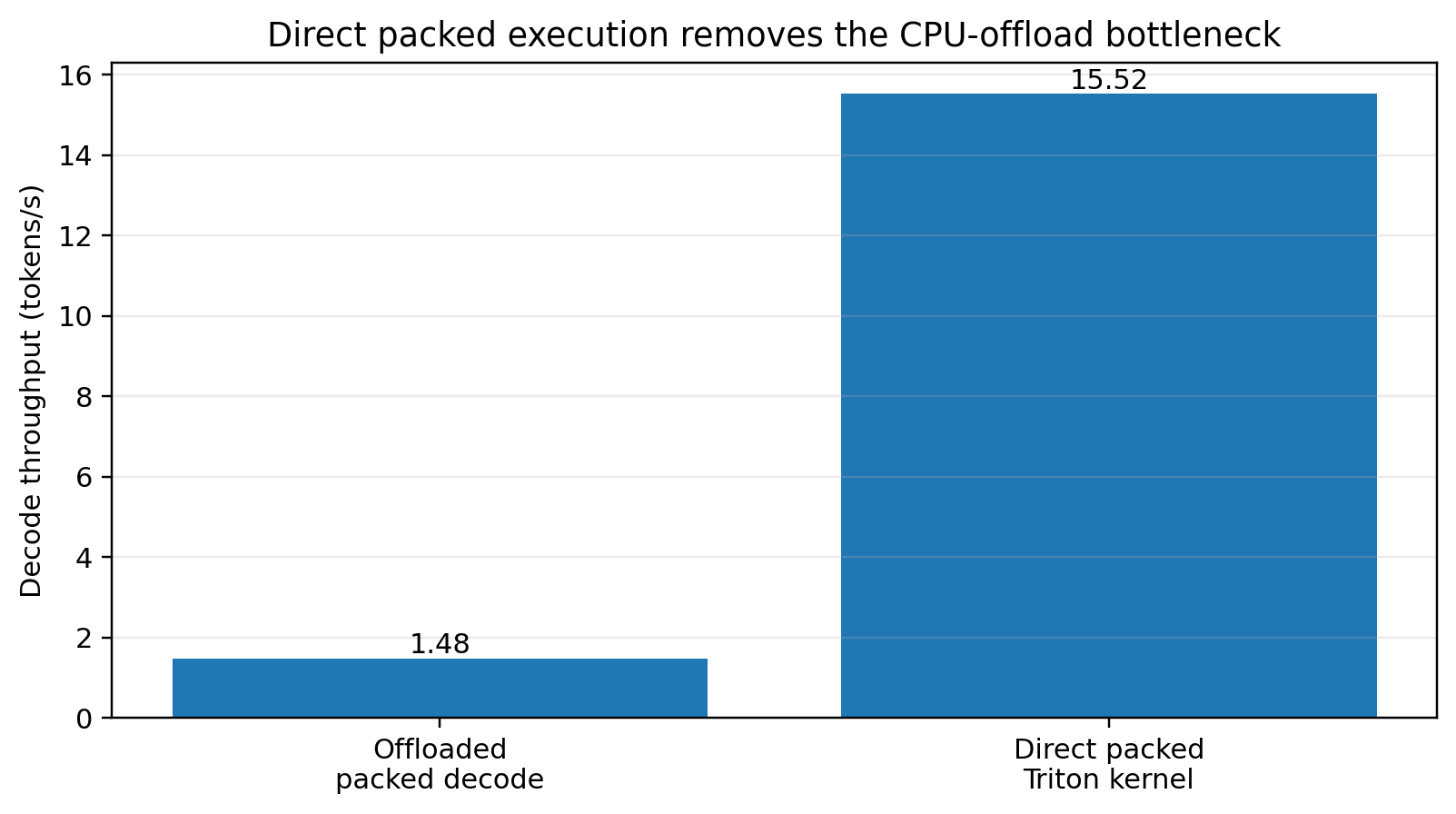}\caption{Execution transition from the offloaded path to direct packed execution. These are different runtime regimes, not a universal FP16 comparison.}\end{figure}

The ratio $15.52/1.48\approx10.49$ shows how strongly the offload bottleneck dominated the earlier result. It should not be reported as a 10.49$\times$ FP16 speedup.

\subsection{Kernel microbenchmark}
At a tested $4096\times2560$ GEMV shape, the measured times were 0.0451 ms for FP16 cuBLAS, 0.2082 ms for the packed Triton kernel, and 0.8924 ms for chunk-unpack plus cuBLAS.

\begin{figure}[H]\centering\includegraphics[width=0.83\linewidth]{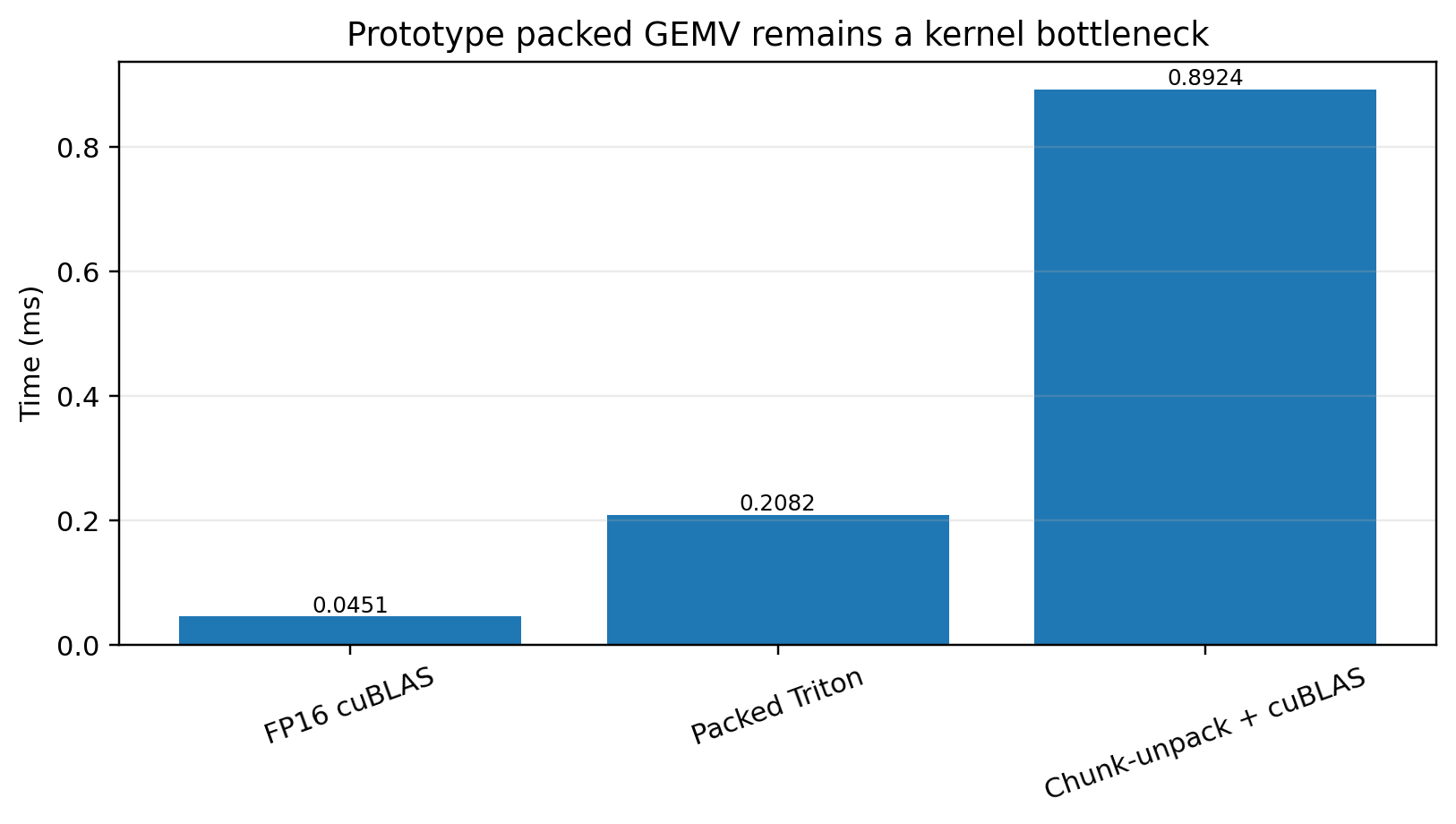}\caption{Prototype packed GEMV timing.}\end{figure}

The packed kernel is therefore 4.6$\times$ slower than the tested FP16 cuBLAS shape. This is a kernel maturity result, not an intrinsic lower bound on ternary arithmetic. The direct end-to-end result demonstrates feasibility; the microbenchmark identifies the optimization gap.

\section{Comparison with Previous OneBit AI Work}
The previous OneBit AI Cloe study investigated a Qwen3.5-0.8B model using QAT and a 72.4M-token recovery budget. Its deployment analysis also showed that disk footprint, runtime memory, load time and throughput must be measured separately.\cite{onebitcloe}

The present study changes the central strategy to PTQ, increases model size to 8B, and adds an explicit external reproduction gate.

\begin{table}[H]\centering
\caption{Evolution across OneBit AI low-bit studies.}
\scriptsize
\begin{tabularx}{\linewidth}{lXXX}\toprule
Dimension & Previous Cloe & Qwen3-4B & Qwen3-8B\\\midrule
Scale & 0.8B & 4B & 8B\\
Conversion & QAT & PTQ & PTQ\\
Recovery/training & 72.4M tokens & Calibration only & Calibration only\\
Primary focus & Capability + learnability & Capability + storage & Validation + scaling + packing + execution\\
Matched retention & -- & 69.6\% & 78.5\%\\
Packed artifact & Prior packed study & 3.96~GiB & 8.24~GiB\\
Direct packed execution & Prior study path & Not central & 15.52 tok/s / 7.35~GiB\\
\bottomrule\end{tabularx}\end{table}

This comparison is not evidence that PTQ universally beats QAT. The models, training budgets and evaluation suites differ. The defensible statement is that the PTQ route now produces a validated 8B low-bit conversion without a large recovery-training run.

\section{Failure Modes and Reproducibility}
The project encountered several failures that could have produced plausible but wrong numbers.

\subsection{Wrong model provenance}
Some 8B scripts initially contained Qwen3-4B identifiers. The resulting apparent ``8B'' perplexity was actually the 4B value. Model identity and quantized checkpoint paths were subsequently made explicit.

\subsection{False ternary artifacts}
A pre-existing artifact looked low-bit but had never been ternarized. Conversely, the real asymmetric model looked non-ternary under zero-count heuristics. The resolution was to require state checks, measurable weight changes and end-to-end PPL reproduction.

\subsection{Memory failures}
A misplaced \texttt{no\_grad} decorator caused graph retention and increased memory from 1.99~GiB to 10.45~GiB over 64 samples. Windows CUDA spill could also create misleadingly large allocations and severe slowdowns. These issues motivated explicit memory instrumentation and a per-process GPU cap.

\subsection{Power loss and resumability}
A reboot near the end of a multi-hour run cost approximately 2.5 hours. Per-layer checkpoints and bit-identical resume reduced future recovery cost to roughly one layer.

\subsection{Benchmark protocol}
Raw accuracy ratios were rejected as retention metrics because chance floors differ. MMLU scoring protocols were explicitly named. Paired model evaluation and provenance assertions became mandatory.

\section{Scientific Interpretation}
The main hypothesis is:
\begin{quote}
\textit{A larger pretrained model can survive aggressive post-training discretization better than a smaller model, while useful deployment requires explicit serialization and direct execution support.}
\end{quote}

The evidence supports the capability part: 78.5\% matched retention and above-chance mathematics. It supports the scaling part in the limited two-point comparison: +8.9 retention points at 8B. It supports the storage part: 16.9~GB to 8.24~GiB with negligible recorded PPL change. It supports the execution-feasibility part: 15.52 tok/s direct packed execution in 7.35~GiB.

It does not yet support a universal speedup claim. The isolated packed GEMV is 4.6$\times$ slower than FP16 cuBLAS, and the FP16 8B model did not fit the same card without offload. The scientifically useful conclusion is that the execution problem has moved from ``can the representation be run directly?'' to ``how far can the kernel be optimized?''

\subsection{Novelty boundary}
We do not claim novelty for ternary weights, the 1.585-bit information limit, KOTMS, E2M-ATQ, GPTQ or the concept of native low-bit inference. The contributions are:
\begin{enumerate}
\item validated 8B scale-up;
\item external reproduction gate;
\item matched 4B/8B capability comparison;
\item effective-bit and representation accounting;
\item lossless lattice-aware packing;
\item direct packed execution;
\item provenance and evaluation safeguards.
\end{enumerate}

\section{Product and Deployment Implications}
\begin{productbox}
\textbf{Compression plus execution is now the strongest product narrative.}
The 8B result is more than a smaller number in a table: the artifact fits in 8.24~GiB, preserves measured perplexity, and can execute directly in 7.35~GiB. This makes local deployment on a 12~GB-class GPU technically plausible. The remaining commercial optimization target is the kernel stack, not the existence of the packed representation.
\end{productbox}

Potential deployment value includes model distribution, local storage, single-GPU feasibility, and hardware co-design. However, no universal serving-cost reduction should be claimed until end-to-end comparisons are repeated across supported hardware and optimized baselines.

The current evidence also suggests a product strategy of selective specialization. The conversion preserves contextual and commonsense behavior more strongly than difficult knowledge retrieval. This does not justify replacing a general FP16 model everywhere, but it motivates evaluation on targeted applications where the retained capability profile matches the workload.

\section{Limitations and Next Experiments}
The principal limitations are:
\begin{enumerate}
\item one seed;
\item $n=500$ capability samples per task;
\item WikiText-2 calibration and evaluation in the reference gate;
\item only two model sizes for the scaling comparison;
\item partial parameter quantization;
\item a prototype rather than production-quality kernel;
\item a 7.9\% local FP16 discrepancy versus the published TWLA baseline;
\item no complete same-protocol comparison against every competing ternary PTQ method.
\end{enumerate}

The highest-value next experiments are multiple seeds, broader calibration such as FineWeb-Edu, direct packed-model capability evaluation, kernel optimization across shapes and batch sizes, embedding/head quantization, additional model sizes, and identical-protocol comparisons against alternative PTQ systems.

\section{Conclusion}
The Qwen3-8B scale-up establishes a strong empirical baseline for aggressive post-training low-bit conversion. The pipeline passes an external reproduction gate, retains 78.5\% chance-corrected capability on the matched suite, and improves retention by 8.9 points over the matched 4B run.

The representation is adaptive rather than a pure single-plane ternary network and has a measured effective weight-information budget near 1.64 bits/weight. Lattice-aware serialization reduces the recorded checkpoint to 8.24~GiB without measurable perplexity cost.

Most importantly for deployment, the project now has a direct packed execution route: 15.52 tokens/s at 7.35~GiB on the tested RTX 5070. The remaining gap is kernel efficiency: the prototype GEMV is 4.6$\times$ slower than FP16 cuBLAS on one shape.

The strongest conclusion is therefore:
\begin{quote}
\textbf{Aggressive post-training conversion can scale into the 8B regime while retaining substantial capability, producing a materially smaller artifact, and enabling direct packed execution. Turning that working path into a broadly efficient serving stack is the next engineering problem.}
\end{quote}

\newpage
\appendix
\section{Experimental Ledger}
\begin{longtable}{p{0.18\linewidth}p{0.26\linewidth}p{0.18\linewidth}p{0.28\linewidth}}
\toprule Stage & Configuration & Outcome & Interpretation\\\midrule
KOTMS & 252 projections & 35 min & Rotation completed\\
E2M-ATQ & A16, 64 samples, $p_d=0.01$ & $\sim$7 h & 8B conversion completed\\
Gate & TWLA reference & 62.00 vs 62.05 & Passed\\
PPL & WikiText-2/C4/PTB & 1.361$\times$ mean & Cross-corpus quality\\
Capability & 8 tasks, $n=500$ & 78.5\% retention & Main capability result\\
Scaling & matched 4B/8B & +8.9 points & De-confounded comparison\\
Packing & lattice-aware & 16.9~GB $\rightarrow$ 8.24~GiB & Lossless re-encoding\\
Packed runtime & direct Triton & 15.52 tok/s, 7.35~GiB & Direct execution\\
Kernel & $4096\times2560$ GEMV & 0.2082 ms & 4.6$\times$ FP16\\
\bottomrule
\end{longtable}

\section{Interpretation Guardrails}
Recommended:
\begin{enumerate}
\item ``The model uses an adaptive ternary representation with an effective budget of approximately 1.64 bits/weight.''
\item ``The 8B model retains 78.5\% chance-corrected capability under the matched eight-task protocol.''
\item ``The matched 8B result is 8.9 retention points above the 4B result.''
\item ``The packed artifact is 8.24~GiB and preserves measured perplexity.''
\item ``Direct packed execution reaches 15.52 tokens/s in 7.35~GiB on the tested RTX 5070.''
\end{enumerate}
Avoid:
\begin{enumerate}
\item ``The whole model is 1.64 bits/parameter.''
\item ``Ternary is universally faster than FP16.''
\item ``The two-point comparison is a scaling law.''
\item ``The method is state of the art'' based on this experiment.
\item ``The local FP16/PPL ratio is directly comparable to the published ratio'' without the baseline caveat.
\end{enumerate}

\section{References}
\bibliographystyle{unsrt}
\bibliography{qwen3_8b_references}
\end{document}